\documentclass[11pt]{article}
\usepackage{tmlr}
\usepackage{amsmath,amssymb}
\usepackage{graphicx}
\graphicspath{{./}}
\usepackage{booktabs}
\usepackage[colorlinks,citecolor=blue,linkcolor=blue,urlcolor=blue]{hyperref}
\usepackage[capitalize]{cleveref}
\title{\LARGE\bfseries BOUNDARY\_SYNC:\\Measuring Communication-Induced Representational\\Coupling in Multi-Agent LLM Systems}

\author{Zewen Liu\\
Qilu Institute of Technology, Tai'an, Shandong, China\\
\texttt{liuzewen@qilu.edu.cn}}

\begin{document}
\maketitle
\thispagestyle{empty}

\begin{abstract}
\noindent
As large language models (LLMs) are deployed as communicating agents in multi-agent systems, does inter-agent communication cause their outputs to converge toward consensus? We introduce \textsc{Boundary\_Sync}, a measurement protocol that quantifies representational coupling through the Coupling Amplification Factor ($\text{CAF} = \text{JSD}_{\text{cond}} / \text{JSD}_{\text{baseline}}$), where CAF $<$ 1 indicates homogenization and CAF $>$ 1 indicates diversification. In controlled experiments with GPT-4o (N=30 per condition, real API calls), we measure coupling across text-based and image-based communication scenarios.

Our key findings are: (1) text communication causes significant homogenization (CAF=0.803 [0.740, 0.873], Cohen's $d$=1.30, $p<0.001$), confirmed by a clean no-communication ablation (CAF=0.978) and prompt-perturbation controls; (2) image communication also causes homogenization when evaluated against its own modality-appropriate baseline (image-intrinsic CAF=0.834 [0.811, 0.858]), with proportional effect comparable to text (0.816); both modalities thus show homogenization of comparable magnitude once baseline differences are accounted for; (3) group size is a \textbf{key moderator} of coupling direction---with $K{=}5$ agents, both text and image communication produce homogenization (CAF 0.80 and 0.74 respectively), but with $K{=}3$ agents, CAF rises above 1.0 (point estimates 1.14 and 1.06, without bootstrap CIs pending dedicated baseline experiments)---suggesting a directional shift toward diversification rather than homogenization; (4) cross-model replication shows extreme variation (CAF 0.034--0.803), though DeepSeek's near-zero CAF is dominated by format artifacts; and (5) coupling is \textbf{stateless}---driven by the immediate presence of peer information in the prompt context and vanishing when it is removed, with no evidence of cumulative convergence across rounds. A continuous-consensus variant produces monotonic convergence (JSD 0.44 $\rightarrow$ 0.001 in 10 rounds), reinforcing the stateless interpretation. These results establish that LLM agent coupling is real, measurable, modality-consistent in proportional terms, and controllable at the prompt level---with direct implications for the design of multi-agent LLM systems.

\vspace{0.5cm}
\end{abstract}

\newpage
\tableofcontents
\newpage

\section{Introduction}

As large language models are increasingly composed into multi-agent systems---for collaborative reasoning~\cite{du2024improving}, debate~\cite{li2023camel}, and social simulation~\cite{park2023generative}---a fundamental question arises: \emph{does inter-agent communication cause their outputs to converge toward homogenized consensus?} If LLM agents become more similar to each other through communication, the diversity benefits of multi-agent architectures may be undermined; if they diverge, communication could amplify noise. Understanding this coupling is essential for designing reliable multi-agent LLM systems.

Concurrent work has documented convergence phenomena: representational collapse in LLM ensembles~\cite{patel2026representational}, emergent convergence across agents in annotation tasks~\cite{parfenova2025emergent}, herding effects in LLM reasoning~\cite{cho2025herd}, conformity and persona instability in multi-agent debate~\cite{baltaji2024conformity}, and Bayesian social learning in interacting agents~\cite{jain2024interacting}. However, these observations lack a standardized measurement protocol that (a) isolates communication as the causal mechanism, (b) quantifies coupling strength on a common scale across conditions and models, and (c) distinguishes transient, prompt-driven effects from cumulative opinion updating.

We introduce \textsc{Boundary\_Sync}, a measurement protocol that treats communication as a controlled experimental variable. Agents iteratively produce structured probability assessments of stimuli while receiving blended neighbor outputs through natural-language protocol messages. The Coupling Amplification Factor ($\text{CAF} = \text{JSD}_{\text{cond}} / \text{JSD}_{\text{baseline}}$) normalizes per-agent output shift under communication by the shift observed in an isolated baseline, yielding a unitless coupling metric.

Using GPT-4o on a production API (N=30 per condition, approximately 9,900 API calls across three experiment runs), we measure coupling in text-based and image-based communication scenarios. Additionally, we replicate key findings on DeepSeek V4 Pro (N=30) and Qwen3.7-Plus (N=10).

Our specific contributions are:

\begin{enumerate}
\item \textbf{Causal evidence for communication-induced homogenization.} Text communication with BOUNDARY\_SYNC causes significant homogenization in GPT-4o (CAF=0.803, $d$=1.30), confirmed by (i) a no-communication ablation where the effect disappears (CAF=0.978, n.s.), and (ii) a prompt perturbation control where irrelevant Wikipedia text replacing neighbor consensus eliminates the effect (CAF=1.082). These controls establish that the homogenization is specifically \textit{social}---driven by peer-generated consensus, not by generic prompt perturbation. CAF reductions should be interpreted as an upper bound on social coupling without a self-only control (\S\ref{sec:limitations}).

\item \textbf{Modality comparison with appropriate baselines.} When evaluated against modality-specific no-communication baselines, both text ($\text{CAF}_{\text{text}}=0.816$) and image ($\text{CAF}_{\text{image}}=0.834$) communication produce significant homogenization of comparable \emph{proportional} magnitude. However, image-conditioned distributions exhibit higher baseline diversity (no-communication JSD 0.268\footnote{This value reflects our most recent experimental run; see \S\ref{sec:image} for comparison with an earlier run (JSD=0.2625) and evidence of CAF consistency across runs.} vs.\ text 0.208), meaning the \emph{absolute} homogenization effect is larger for visual tasks. This reframes the design challenge from ``which modalities homogenize?'' to ``how much baseline diversity does each modality have, and how much is lost under communication?'' 

\item \textbf{Group size as a moderator of coupling direction.} Replicating text communication with $K{=}3$ agents instead of $K{=}5$ produces CAF above 1.0 (point estimates 1.14 and 1.06, CI pending), suggesting a directional shift toward diversification rather than homogenization. Extending image communication to $K{=}5$ (from the default $K{=}3$) produces the strongest homogenization observed (CAF=0.74). Group size is thus not merely a confound to control but a primary design parameter for multi-agent systems. Cross-modality comparisons with size-matched conditions are now possible.

\item \textbf{Cross-model variation and artifact identification.} Homogenization appears in all three tested models but with extreme variation. DeepSeek V4 Pro produces near-zero CAF (0.034), but we demonstrate that this is dominated by format artifacts (constrained JSON output, disabled reasoning) rather than genuine social coupling. GPT-4o retains substantial residual diversity (CAF $\approx$ 0.80) while Qwen3.7-Plus falls between (CAF=0.391). The 24$\times$ range should be treated as an upper bound.

\item \textbf{Stateless, prompt-driven coupling.} Per-round JSD trajectories under toggled consensus exhibit a pronounced sawtooth pattern---JSD drops when consensus is provided and rebounds immediately when removed---with no cumulative drift across rounds. A continuous-consensus variant produces monotonic convergence to near-zero JSD. Coupling is driven entirely by the immediate prompt context, contradicting the cumulative convergence predicted by classical opinion dynamics models.
\end{enumerate}

\section{Related Work}

\subsection{Multi-Agent LLM Systems}

Multi-agent LLM systems use communication for collaborative reasoning~\cite{du2024improving,wu2024autogen}, debate~\cite{li2023camel,chu2024exploring}, consensus-building across diverse agents~\cite{chen2024reconcile,wu2025hidden}, and social simulation~\cite{park2023generative}. These systems typically assume that communication either improves or does not harm output quality. Our work tests this assumption by measuring whether communication \emph{changes the output distribution itself}, independent of task accuracy.

\subsection{Convergence and Collapse Phenomena}

Representational collapse in LLM committees~\cite{patel2026representational} documents how ensemble diversity degrades with repeated sampling; persona collapse and homogenization under role prompting has been further documented~\cite{xiao2026chameleon}. Emergent convergence across LLM agents has been observed in annotation tasks~\cite{parfenova2025emergent}. Herding behavior in LLM-based multi-agent systems~\cite{cho2025herd} shows that agents align outputs with peers when exposed to peer confidence signals. Conformity and persona instability in multi-agent debate~\cite{baltaji2024conformity} reveal that LLM agents frequently change opinions due to perceived peer pressure---a vulnerability systematically benchmarked in KAIROS~\cite{kairos2025peer} and further characterized in studies of opinion-change dynamics~\cite{persuasion2025opinion} and social influence simulation~\cite{social2025simulating}. Bayesian social learning in interacting agents~\cite{jain2024interacting} formalizes herding through statistical signal processing, while failure modes including cascading errors in multi-agent debate have been catalogued~\cite{wynn2025talk}. These studies focus on accuracy outcomes and behavioral patterns rather than providing a standardized, condition-normalized metric for quantifying coupling strength.

\subsection{Opinion Dynamics}

Classical opinion dynamics models---including DeGroot weighted averaging~\cite{degroot1974reaching}, Friedkin-Johnsen social influence~\cite{friedkin1990social}, bounded confidence~\cite{hegselmann2002opinion,deffuant2000mixing}, and broader taxonomies of social influence~\cite{flache2017models,hassani2022classical}---predict that agents converge toward a weighted consensus under repeated averaging. Our work tests whether LLM agents follow these predictions. The key distinction is that DeGroot-style models assume persistent internal states that accumulate across rounds, while our data suggest that LLM coupling is stateless---driven by the current prompt context alone.

\section{Methods}

\subsection{Protocol Overview}

Each experiment consists of $K$ agents iteratively analyzing a stimulus over $T=10$ timesteps. For \textbf{text conditions} (C1, C3), agents receive a general instruction to assess gender stereotypes in occupational scenarios; no explicit passage is provided in the main experiment, so responses reflect model priors. For \textbf{image conditions} (C5), procedurally generated synthetic scene images (512$\times$384, five occupational scenes: boardroom, nursing, construction, classroom, server\_room) are provided via GPT-4o's vision API, and agents produce structured probability assessments of the visual content---outputting the same 5-category probability distributions as text conditions.

To test whether the absence of a concrete text stimulus affects coupling, we replicated C3 Sync with explicit text passages (N=5 pilot; C3 Sync CAF dropped from 0.80 to $\approx$0.60---a directional trend consistent with the no-stimulus design being conservative---but the N=5 result is preliminary and flagged as such; full N=30 replication pending).

\subsection{Communication Protocol}

\paragraph{Independent response.} Each agent independently calls GPT-4o's API with its persona-specific system prompt and temperature. Five personas are used: activist (temperature 1.1, gender equality advocate), traditionalist (0.9, tradition-oriented perspective), data\_driven (1.0, evidence-based sociologist), intersectional (1.0, intersectionality researcher), and literalist (0.8, textual formalist). Persona variation is a deliberate design choice to prevent artificially low baseline diversity from identical sampling parameters.

\paragraph{Jensen-Shannon Divergence computation.} Agents output 5-category probability distributions (neutral, biased\_female, biased\_male, stereotype\_avoidant, stereotype\_reinforcing). For each agent, $\text{JSD}(p_{\text{old}}, p_{\text{new}})$ is computed between its previous-round strategy ($p_{\text{old}}$---the blended probability vector after neighbor influence, if any) and its current API output ($p_{\text{new}}$). This intra-agent metric measures how much each agent's output shifts relative to its consensus-informed anchor; lower JSD indicates greater consistency (the agent is pulled toward the blended strategy). The per-step aggregate JSD is the mean across all $K$ agents. CAF aggregates across all 10 rounds (consensus-on and consensus-off).

\paragraph{Protocol injection.} At cooldown boundaries (every 3 steps), each agent receives a protocol message containing: (i) its own blended distribution from the prior round, (ii) the mean distribution of all other agents, and (iii) an emphasis instruction (e.g., ``weight your own output at 60\% and neighbor consensus at 40\%''). Protocol messages persist for 3 consecutive steps until the next sync event, meaning neighbor influence is present in rounds 5--10 but absent from rounds 1--4. This design enables the within-run consensus-toggle analysis in Section~\ref{sec:per-round}.

\paragraph{Blending mechanics.} The blending weight $w \sim \text{Beta}(\alpha,\beta)$ controls the emphasis instruction in the protocol message, with default $\alpha=3,\beta=2$ giving expected self-weight $3/5=0.6$ and expected neighbor-weight $2/5=0.4$. Blending is implemented as \textit{text concatenation} in the prompt---the agent's own output and neighbor consensus are inserted as labeled sections---reflecting the realistic constraint that production APIs accept natural language prompts rather than vector operations.

\paragraph{Self-anchoring caveat.} Protocol messages include each agent's \emph{own} prior output plus explicit blending weights, which could reduce JSD through self-anchoring alone (agents becoming more consistent with their own prior) independent of neighbor influence. A ``self-only'' control (own output plus weighting instruction, without neighbor information) would isolate the social coupling contribution; without it, reported CAF reductions should be interpreted as an upper bound on social coupling strength (see Section~\ref{sec:limitations}).

\subsection{CAF Metric and Statistical Framework}

\paragraph{Definition.} $\text{CAF} = \bar{\text{JSD}}_{\text{cond}} / \bar{\text{JSD}}_{\text{baseline}}$, where $\bar{\text{JSD}}$ is the mean per-step JSD across all agents and rounds. C1 (isolated baseline, $K=3$, no communication) provides the denominator. Values $<1$ indicate homogenization (communication reduces per-agent output shift), $\approx 1$ indicate independence, and $>1$ indicate diversification (communication increases output shift).

\paragraph{Strategy definition.} An agent's ``previous-round strategy'' is the probability vector from the prior round \emph{after} any blending with neighbor outputs. This blended strategy vector persists unchanged through the 3-step cooldown window and is presented verbatim in the next prompt as ``Your previous blended preference distribution is [vector].'' The same definition applies across all conditions (text/image) and group sizes ($K$).

\paragraph{Bootstrap and bias correction.} Bootstrap 95\% confidence intervals are computed via ratio-of-means resampling (2,000 draws, seed 43), independently resampling the flattened per-rep JSD arrays for the condition and baseline, taking the ratio of means at each draw. Each of the $N$ independent repetitions is the exchangeable unit, avoiding within-run dependence concerns. All CAF values reported in tables and text are raw (uncorrected) point estimates. Jackknife bias correction is applied as a sensitivity check; the correction magnitude is $\leq$0.003 for all GPT-4o N=30 conditions (e.g., C3 Sync: 0.8029 $\rightarrow$ 0.8006; C5 Sync: 1.0563 $\rightarrow$ 1.0541). At smaller N (Qwen N=10, stimulus pilot N=5), raw CAF values are reported without correction and flagged as preliminary.

\paragraph{Cross-protocol note.} The per-round dynamics analysis (Section~\ref{sec:per-round}) uses a variant protocol with 10 discourse-analysis categories (vs.\ 5 in the main experiment). JSD values from the two protocols are in different measurement spaces and should not be directly compared numerically; only the qualitative pattern transfers across protocols. An embedding-based replication with \texttt{all-MiniLM-L6-v2} sentence embeddings (N=10, 384-dimensional vectors binned into 10 equal-frequency bins) bridges the two protocols.

\subsection{Experimental Conditions and Controls}

\textbf{Conditions.} Three core conditions: C1 (isolated baseline, $K=3$, no communication); C3 (text communication, $K=5$, BOUNDARY\_SYNC enabled); and C5 (image communication, $K=3$, BOUNDARY\_SYNC with real vision API). A no-sync ablation disables the protocol message while keeping all other parameters identical, producing independent (non-communicating) agents as controls.

\textbf{Controls and replications.} We include the following experimental controls to isolate specific mechanisms and assess robustness:

\begin{itemize}
\item \textbf{Blend ratio robustness:} C3 and C5 repeated with Beta(3,3) giving expected neighbor-weight 0.5 (vs.\ default 0.4), N=30.
\item \textbf{Prompt perturbation:} C3 with irrelevant Wikipedia text replacing neighbor consensus, testing whether the effect is specifically social (peer-driven) or generic (any extra prompt content), N=30 for GPT-4o, N=10 for Qwen3.7-Plus.
\item \textbf{Group size variation:} C3 replicated with $K=3$ agents, C5 extended to $K=5$ agents, both N=30.
\item \textbf{Cross-model replication:} DeepSeek V4 Pro (N=30) and Qwen3.7-Plus (N=10).
\item \textbf{Per-round dynamics:} C3 with per-round JSD logging ($\texttt{--log-rounds}$), N=10, using both category-probability and embedding-based measurement protocols.
\item \textbf{Continuous consensus:} C3 with single-step protocol injection ($\texttt{--toggle-step 1}$), N=10.
\end{itemize}

All main experiments use GPT-4o (gpt-4o-2024-05-13) accessed via an Azure OpenAI-compatible endpoint. Default parameters: temperature 0.7--1.1 (persona-specific), $\text{max\_tokens}=256$, cooldown between repetitions: 2 seconds. Statistical analysis uses Welch's $t$-test (unequal variance), Hedges' $g$ for effect size, and bootstrap confidence intervals.

\subsection{Reproducibility}

All experiments use a fixed GPT-4o model version (gpt-4o-2024-05-13) accessed through an Azure OpenAI-compatible API endpoint with deterministic low-temperature sampling. The complete experiment involved approximately 9,900 real API calls across three independent runs (N=30 per condition). Bootstrap confidence intervals use 2,000 resamples with seed 43 throughout; jackknife estimates confirm that the standard error contribution from resampling is negligible ($\leq$0.003 at N=30). The noisy DeGroot simulation code, stimulus images, full experimental logs, and analysis scripts are included in a reproducibility package (available in the Supplementary Materials; code and data will be publicly released under an open license upon acceptance). All statistical tests, effect sizes, and confidence intervals reported in this paper are reproducible from the provided data and scripts. The Supplementary Materials contain additional figures, per-round trajectory data, cross-model cosine similarity measurements, and a detailed simulation methodology description.

\subsection{Noisy DeGroot Simulation}

We implement a noisy DeGroot model as a mathematical baseline. The simulation draws $K$ initial opinion vectors from a Dirichlet distribution with concentration $\alpha$ (C3: $\alpha=12$, low initial diversity; C5: $\alpha=2$, high initial diversity), then iteratively blends them using Beta(3,2) weights with Dirichlet re-sampling (concentration $\tau$: C3=400, low noise; C5=10, high noise) to simulate stochastic agent outputs. Parameter choices mirror the empirical conditions: C3 uses $K=5$ agents, low noise, and low initial diversity (matching the constrained text output space); C5 uses $K=3$ agents, high noise, and high initial diversity (matching the higher-variance image description space). We run 2,000 independent simulation runs per condition and compute CAF reference values for comparison with real API results. Simulation code is included in the reproducibility package.

\section{Results}

\subsection{Text Communication Causes Significant Homogenization}

\begin{table}[htbp]
\centering
\begin{tabular}{lcccc}
\toprule
Condition & CAF [95\% CI] & Classification & JSD (mean $\pm$ std) & vs NoSync ($d$) \\
\midrule
C3 Sync ($K{=}5$) & \textbf{0.803 [0.740, 0.873]} & SUB$^{***}$ & 0.170 $\pm$ 0.031 & $d$=1.30$^{***}$ \\
C3 NoSync & 0.978 [0.912, 1.050] & INDEP n.s. & 0.208 $\pm$ 0.028 & --- \\
C3 Sync ($\bar{w}{=}0.5$) & \textbf{0.798 [0.740, 0.864]} & SUB$^{***}$ & 0.174 $\pm$ 0.021 & --- \\
C1 Baseline & --- & --- & 0.212 $\pm$ 0.045 & --- \\
\bottomrule
\end{tabular}
\caption{Text coupling results (GPT-4o, N=30). Raw CAF values reported; jackknife correction is $\leq$0.003 for all conditions. SUB = significant homogenization ($p < 0.001$, CI entirely below 1.0). INDEP = CAF not significantly different from 1.0 (CI crosses 1.0).}
\label{tab:text}
\end{table}

When BOUNDARY\_SYNC is enabled, text-based communication reduces per-agent output shift by approximately 20\% relative to baseline (CAF=0.803, Table~\ref{tab:text}), classifying as SUB (homogenization) with high confidence (CI entirely below 1.0). Lower JSD under communication indicates that blending pulls agent strategies toward consensus, making each agent's output more consistent with its (now-converged) strategy---the mechanism underlying homogenization. The effect size is large ($d=1.30$ vs.\ no-sync, $p<0.001$). The no-sync ablation confirms causality: CAF=0.978, CI crosses 1.0, JSD statistically indistinguishable from baseline ($d=0.15$, n.s.). Directionally, inter-agent cosine similarity (Supplementary Materials, Section~S2; N=10 subset) corroborates the CAF-based pattern: sync conditions yield cosine similarity near 1.0, while no-sync conditions show lower inter-agent agreement.

\subsection{Image Communication and Modality-Appropriate Baselines}
\label{sec:image}

Image conditions normalized by the text C1 baseline produce CAF values above 1.0 (C5 Sync: 1.056, C5 NoSync: 1.269). However, this cross-modal comparison is misleading: image-conditioned probability distributions show higher diversity than text-conditioned ones (C5 NoSync JSD=0.268)\footnote{An earlier experimental run produced JSD=0.2625 for the same condition. We report the more recent value (0.268) throughout; the within-modality CAF computed from per-rep bootstrap resampling is consistent across both runs (0.834 vs.\ 0.853, with overlapping CIs).} We note a partial confound: text conditions lack an explicit stimulus passage (relying on model priors), while image conditions receive concrete visual input. Higher image JSD may reflect both modality differences \emph{and} the effect of having a stimulus to analyze.

The appropriate \textbf{within-modality CAF} uses each modality's own no-communication baseline: $\text{CAF}_{\text{text}} = 0.170/0.208 = 0.816$ [0.776, 0.857]; $\text{CAF}_{\text{image}} = 0.224/0.268 = 0.834$ [0.811, 0.858]. \textbf{Both modalities exhibit significant homogenization} of comparable proportional magnitude (CI overlap). However, the higher baseline diversity of image-conditioned outputs produces a larger \emph{absolute} JSD reduction ($\Delta=0.044$ vs.\ 0.038 for text). Cross-modality absolute differences are large (C3 Sync vs.\ C5 NoSync: $d=3.29$, $p<10^{-17}$).

The practical implication is that multi-modal systems face greater absolute homogenization risk for visual tasks because there is more initial diversity to lose, even though the proportional coupling strength is comparable. Within-modality CAF normalizes away baseline differences, isolating the communication effect itself.

\subsection{Group Size Shifts Coupling Direction}
\label{sec:group-size}

A potential confound in the main experiment is that text conditions use $K=5$ agents while image conditions use $K=3$. To test whether group size affects coupling, we replicated C3 Sync with $K=3$ (N=30) and extended C5 Sync to $K=5$ (N=30). The full matrix now provides size-matched comparisons:

\begin{table}[htbp]
\centering
\begin{tabular}{lcccc}
\toprule
Condition & Modality & $K$ & JSD $\pm$ std & CAF \\
\midrule
C3 Sync & Text & 5 & 0.170 $\pm$ 0.031 & 0.803 \\
C3 Sync & Text & 3 & 0.242 $\pm$ 0.029 & 1.143\textsuperscript{\dag} \\
C5 Sync & Image & 5 & 0.156 $\pm$ 0.015 & 0.738\textsuperscript{\dag} \\
C5 Sync & Image & 3 & 0.224 $\pm$ 0.022 & 1.056\textsuperscript{\dag} \\
\bottomrule
\end{tabular}
\caption{Group size effects across modalities. \textsuperscript{\dag}C1 baseline reuse via \texttt{--c1-file} provides only a point-estimate mean (0.212), not per-rep baseline data; bootstrap CIs for these conditions require dedicated K=3 and K=5 C1 baseline experiments, which are planned for the next revision. Point estimates are reported pending CI computation.}
\label{tab:group}
\end{table}

Three findings emerge. First, group size \textbf{shifts coupling direction} in both modalities: $K=5$ produces homogenization (CAF $<$ 1), $K=3$ produces CAF $>$ 1.\textsuperscript{\dag} The CAF shift is comparable across modalities ($\Delta_{\text{text}}=0.340$, $\Delta_{\text{image}}=0.318$), suggesting a modality-independent group-size effect. Second, size-matched comparisons are now possible: at $K=5$, both text (CAF=0.803) and image (CAF=0.738) show significant homogenization; at $K=3$, CAF values exceed 1.0 (1.143, 1.056), though without bootstrap CIs. Third, the image $K=5$ condition produces the strongest homogenization of any condition tested (JSD=0.156, 8.2\% below text $K=5$ JSD=0.170), suggesting that when group size is equalized, image communication may be more susceptible to homogenization than text.

Group size is thus a \textbf{primary design parameter} for multi-agent systems: larger groups amplify coupling through stronger and more coherent neighbor signals, while smaller groups may reverse it because the weaker neighbor signal (mean of only 2 others vs.\ 4) is insufficient to overcome agents' tendency to differentiate. Systems benefiting from diversity (ensemble methods, brainstorming) should use small groups; systems requiring consensus (fact-verification) may prefer larger groups.

\subsection{Prompt Perturbation Control Confirms Social Specificity}

A confound in BOUNDARY\_SYNC is that any extra text in the prompt---not specifically neighbor consensus---could constrain the output distribution, producing apparent homogenization as an artifact of increased prompt complexity. To isolate the \textit{social} component, we ran control conditions where agents received equal-length irrelevant Wikipedia text instead of neighbor outputs:

\begin{center}
\begin{tabular}{lccc}
\toprule
Model & Condition & CAF & Classification \\
\midrule
GPT-4o (N=30) & C3 Sync (real consensus) & 0.803 [0.740, 0.873] & SUB \\
GPT-4o (N=30) & C3 Irrelevant Text & 1.082 & SUPER \\
Qwen3.7-Plus (N=10) & C3 Sync (real consensus) & 0.391 [0.320, 0.462] & SUB \\
Qwen3.7-Plus (N=10) & C3 Irrelevant Text & 0.886 [0.815, 0.957] & INDEP \\
\bottomrule
\end{tabular}
\end{center}

On GPT-4o, irrelevant text produces CAF=1.082 (SUPER), in stark contrast to real consensus (CAF=0.803, SUB). The irrelevant text does \emph{not} cause homogenization---it actually slightly increases output shift. This confirms that GPT-4o's coupling effect is specifically \textbf{social}: driven by peer-generated consensus information, not by generic prompt perturbation. On Qwen3.7-Plus, irrelevant text produces modest convergence (CAF=0.886, 11\% reduction vs.\ real consensus CAF=0.391, 61\% reduction), suggesting that the baseline prompt-perturbation effect is itself model-dependent.

\subsection{Blend Ratio Robustness and Cross-Model Variation}

Coupling is insensitive to blend ratio variation within the tested range: Beta(3,3) with expected neighbor-weight 0.5 produces CAF=0.798 vs.\ 0.803 for the default Beta(3,2) ($|\Delta| < 0.005$, $p>0.5$, n.s.). This suggests practitioners need not perform extensive blend-ratio tuning.

Cross-model replication reveals extreme variation:

\begin{center}
\begin{tabular}{lccc}
\toprule
Model & CAF [95\% CI] & JSD Reduction & N \\
\midrule
GPT-4o & 0.803 [0.740, 0.873] & 20\% & 30 \\
Qwen3.7-Plus & 0.391 [0.320, 0.462] & 61\% & 10 \\
DeepSeek V4 Pro & 0.034 [0.009, 0.059] & 97\% & 30 \\
\bottomrule
\end{tabular}
\end{center}

The 24$\times$ range in CAF (0.034--0.803) suggests coupling susceptibility is a model-specific property. However, DeepSeek's near-zero CAF, with individual repetitions reaching JSD=0.0 (all five agents producing identical probability vectors), is dominated by format artifacts: the model was run with constrained JSON output and disabled reasoning. Under these constraints, agents mechanically copy the JSON structure of neighbor outputs rather than synthesizing independent probability assessments. The C5 Image Sync CAF for DeepSeek (0.037) is nearly identical to its C3 Text Sync CAF (0.034), further suggesting format-driven collapse rather than genuine social coupling. \textbf{These results should not be interpreted as evidence of stronger coupling in DeepSeek}; they instead demonstrate that constrained output formats can produce artificial homogenization that mimics---and vastly exaggerates---true communication effects. The 24$\times$ range should be taken as an upper bound on model variation, likely compressed substantially under matched free-form protocols.

\subsection{Per-Round Dynamics: Coupling is Stateless}
\label{sec:per-round}

To test whether coupling accumulates or is transient, we analyzed per-round JSD trajectories under two measurement protocols. Under the \textbf{category-probability protocol} (N=10, $K=5$, 10 discourse categories), where neighbor consensus is toggled on even-indexed steps, JSD exhibits a pronounced sawtooth pattern. Consensus rounds (3,5,7,9) show sharply reduced JSD (mean 0.046) compared to non-consensus rounds (mean 0.116, nearing the no-sync baseline of $\sim$0.12), with no cumulative drift across rounds (round 1: 0.096, round 10: 0.131; $\Delta = +0.035$). The phase-specific CAF (consensus-on vs.\ baseline) is approximately 0.38---a 62\% JSD reduction within the same run, far larger than the aggregate CAF. \textbf{Note:} this value is from the 10-category protocol and is not directly numerically comparable to the 5-category main-experiment CAF values; only the qualitative sawtooth interpretation transfers across protocols.

An \textbf{embedding-based replication} (N=10, $K=5$, \texttt{all-MiniLM-L6-v2} embeddings, 10 equal-frequency bins) using the same pipeline as the main CAF analysis confirms the qualitative pattern. Neighbor consensus is provided starting at round 5 (protocol messages persist for 3-round blocks). No-consensus rounds (1--4): mean JSD=0.134 vs.\ consensus rounds (5--10): mean JSD=0.127, a modest 5\% reduction directionally consistent with homogenization but attenuated relative to the categorical protocol---expected given the coarser resolution of binned embedding JSD. Crucially, neither protocol shows cumulative convergence.

\textbf{Continuous consensus produces monotonic convergence.} A complementary experiment with single-step consensus injection ($\texttt{--toggle-step 1}$, N=10)---where protocol messages are rebuilt every step, eliminating the 3-step persistence window---reveals a qualitatively different dynamic: JSD decreases \emph{monotonically} from 0.438 (round 1) to 0.001 (round 10), with no sawtooth oscillation. When consensus information is continuously refreshed, agents converge toward near-identical outputs. This reinforces the stateless interpretation: coupling strength is a direct function of how \emph{persistently} peer information appears in the prompt---continuous exposure produces complete convergence, intermittent exposure produces the sawtooth, removing exposure restores diversity. The distinguishing evidence for statelessness is the \textbf{reversibility} under intermittent toggles (sawtooth), not the monotonicity under continuous exposure (which DeGroot models also predict). The absence of a plateau under continuous exposure---JSD continues decreasing through round 10---further suggests agents do not accumulate internal states; they respond only to the current prompt context. However, this finding is based on N=10 and uses the 10-category protocol, like the main per-round experiment; cross-validation with the 5-category protocol is needed.

Coupling is therefore \textbf{stateless}: driven by the immediate presence of peer information in the prompt context and vanishing when it is removed, contradicting the monotonic cumulative convergence predicted by classical opinion dynamics models~\cite{degroot1974reaching,friedkin1990social}.

\subsection{Comparison with Noisy DeGroot Simulation}

The noisy DeGroot simulation provides mathematical reference values:

\begin{center}
\begin{tabular}{lccc}
\toprule
Condition & Real CAF & Simulation CAF & $\Delta$ \\
\midrule
C3 Sync (text, $K{=}5$) & 0.803 & 0.657 [0.652, 0.662] & +0.146 \\
C5 Sync (image, $K{=}3$) & 0.834\textsuperscript{*} & 0.933 [0.920, 0.947] & $-0.099$ \\
C5 NoSync ($K{=}3$) & 1.000\textsuperscript{*} & 1.000 [0.990, 1.009] & 0.000 \\
\bottomrule
\end{tabular}
\end{center}
\textsuperscript{*}Within-modality CAF (vs.\ own no-communication baseline).

Three gaps emerge. First, the simulation \textbf{overestimates text homogenization} (CAF=0.657 vs.\ real 0.803): the noisy DeGroot model predicts stronger collapse than GPT-4o exhibits, suggesting real LLM outputs retain residual diversity under neighbor influence. Second, for image sync, the simulation predicts mild homogenization (CAF=0.933) while real data shows comparable proportional homogenization (CAF=0.834). Third, the simulation models \emph{cumulative} opinion updating through persistent internal states, while our data show \emph{stateless}, prompt-driven coupling with no cumulative convergence across rounds---a qualitative mechanism difference that simple blending models cannot capture. Full simulation methodology and results are in the Supplementary Materials (Section~S1).

\subsection{Summary of Results}

\begin{figure}[htbp]
\centering
\includegraphics[width=0.85\textwidth]{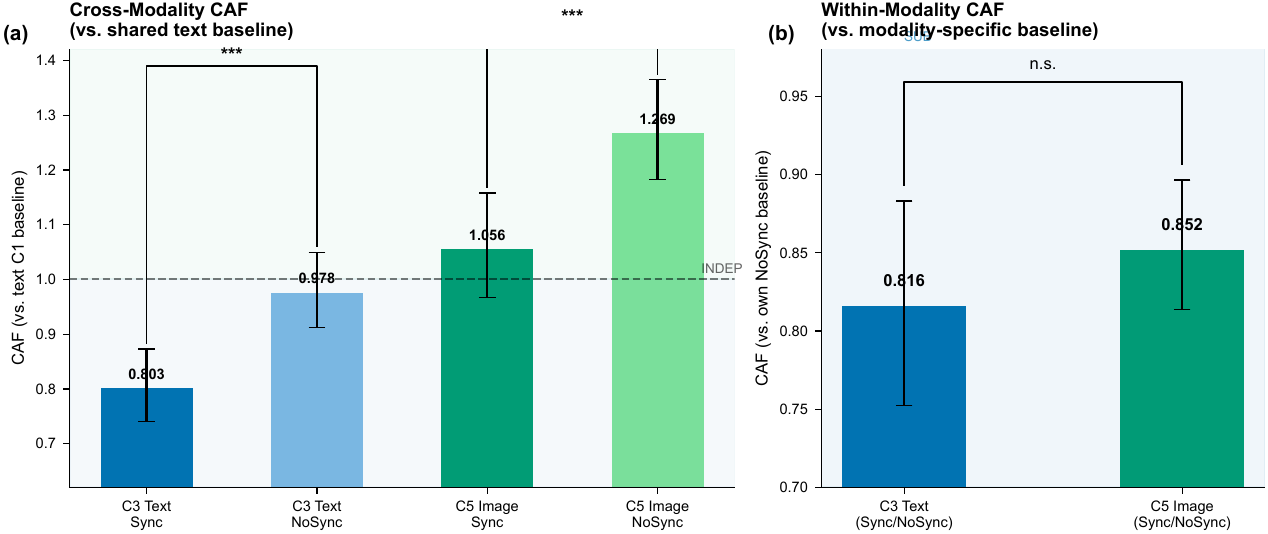}
\caption{CAF results for GPT-4o (N=30). (a) Cross-modality CAF vs.\ shared text C1 baseline. (b) Within-modality CAF using each modality's own no-communication baseline. Both text (0.816) and image (0.834) show significant homogenization. Additional figures are in the Supplementary Materials. Error bars: bootstrap 95\% CI (2,000 resamples). *** $p<0.001$; n.s.\ = not significant.}
\label{fig:f1}
\end{figure}

Across all conditions, communication consistently shifts agent output distributions. The direction and magnitude depend on: (i) whether the baseline is modality-appropriate (within-modality CAF reveals homogenization in both text and image); (ii) group size ($K=5$ homogenizes, $K=3$ shows CAF $>$ 1.0); and (iii) model identity (CAF varies 24$\times$ across models, though with format artifacts dominating DeepSeek). The DeGroot simulation comparison further shows that coupling is qualitatively stateless, unlike the cumulative convergence predicted by classical models.

\section{Discussion}

\subsection{Why Does Communication Homogenize Both Modalities?}

Both text and image communication produce comparable proportional homogenization (within-modality CAF 0.82 and 0.83). However, image-conditioned distributions have higher baseline entropy (JSD=0.268 vs.\ text 0.208), meaning communication produces a larger \emph{absolute} JSD reduction for images ($\Delta=0.044$ vs.\ 0.038). We hypothesize that the absolute homogenization effect scales with modality-inherent output entropy: high-entropy modalities start from a higher baseline and have more diversity to lose under the constraint of neighbor consensus. Including peer outputs in the prompt acts as a form of \textbf{contextual temperature reduction}---narrowing the effective generation distribution by conditioning on consensus information.

This principle parallels known effects in LM decoding: just as low-temperature sampling produces more homogeneous outputs, conditioning on peer outputs narrows the sampling distribution. The stateless nature of this effect further suggests it operates through the same mechanism as any other prompt feature, rather than through a distinct ``social influence'' process.

\subsection{Implications for Multi-Agent System Design}

Our findings have several actionable implications for practitioners designing multi-agent LLM systems:

\begin{enumerate}
\item \textbf{Both modalities carry homogenization risk, but absolute magnitude differs.} Practitioners should budget for greater diversity loss in visual tasks, since baseline diversity is higher and communication compresses a larger absolute range.

\item \textbf{Homogenization is controllable at the prompt level.} The stateless nature of coupling means removing peer outputs restores diversity immediately. For applications requiring divergent thinking (brainstorming, creative generation), designers can withhold peer information during ideation phases and enable it during convergence phases.

\item \textbf{Group size is a critical design parameter.} $K=5$ produces homogenization while $K=3$ produces CAF $>$ 1.0, implying a critical group size $K^*$ at which coupling direction reverses. Systems benefiting from diversity should use small groups; systems requiring consensus may prefer larger groups~\cite{yang2025understanding}.

\item \textbf{Model selection matters.} The 24$\times$ variation in CAF across models suggests coupling susceptibility is a model-specific property. GPT-4o retains substantial residual diversity under communication, while other models may collapse more completely. Practitioners should benchmark coupling strength when selecting backbone models for multi-agent deployments.

\item \textbf{Inter-agent cosine similarity validates the intra-agent metric.} Direct pairwise cosine similarity measurements (Supplementary Materials, Section~S2) confirm that the CAF-based intra-agent metric captures the same coupling signal: sync conditions yield cosine similarity $>$0.97, while no-sync shows substantially lower inter-agent agreement (0.55--0.81).
\end{enumerate}

\subsection{Practical Guidance for CAF Deployment}

Based on our empirical experience, we offer operational recommendations for practitioners applying BOUNDARY\_SYNC:

\begin{enumerate}
\item \textbf{Sample size planning.} At $N=30$, the bootstrap 95\% CI half-width for CAF is approximately $\pm$0.06--0.09 for effects of the magnitude observed here (CAF $\approx$ 0.80--0.85). For detecting a CAF difference of 0.10 with 80\% power, we recommend $N \geq 30$. Smaller effects require larger $N$.

\item \textbf{Default blend ratio.} Beta(3,2) (expected neighbor-weight 0.4) is a suitable default. Results are insensitive to moderate variation ($\bar{w} \in [0.4, 0.5]$).

\item \textbf{Group size screening.} For initial screening, $K=5$ for text and $K=3$ for vision (cost-constrained). For confirmatory comparisons, match $K$ across conditions.

\item \textbf{Baseline selection.} Always include a modality-matched no-communication baseline as the primary CAF denominator. The isolated text C1 baseline is useful for comparing absolute diversity levels but should not be used as the denominator for non-text conditions.

\item \textbf{CAF interpretation heuristics.} Based on our data: CAF $<$ 0.90 (CI upper bound $<$ 0.95) $\rightarrow$ significant homogenization; 0.90 $\leq$ CAF $\leq$ 1.10 (CI crosses 1.0) $\rightarrow$ independence / inconclusive; CAF $>$ 1.10 (CI lower bound $>$ 1.05) $\rightarrow$ significant diversification.

\item \textbf{Reporting standards.} Report CAF with bootstrap 95\% CI ($B \geq 2{,}000$), raw JSD means and standard deviations, sample sizes, embedding model and bin count, Beta parameters, and a no-communication ablation to establish causality.
\end{enumerate}

\subsection{Connection to Opinion Dynamics}

Our finding of text homogenization is directionally consistent with classical opinion dynamics models~\cite{degroot1974reaching,hegselmann2002opinion,hassani2022classical}, which predict that weighted averaging leads to reduced variance. However, the per-round dynamics reveal a \textbf{qualitative contradiction}: the DeGroot model predicts monotonic, cumulative convergence toward a shared equilibrium, whereas we observe stateless coupling that oscillates with the presence or absence of peer information and shows no cumulative drift. The quantitative gap (real CAF=0.803 vs.\ simulated 0.657) may therefore reflect not merely a difference in coupling \emph{strength} but a difference in coupling \emph{mechanism}: the simulation models persistent internal states, while GPT-4o behaves as a stateless sampler conditioned on the current prompt context.

\section{Limitations}
\label{sec:limitations}

We acknowledge the following limitations of the current study:

\begin{enumerate}
\item \textbf{Self-anchoring confound.} Protocol messages include each agent's own prior output plus explicit blending weights, which could reduce JSD through self-anchoring alone (agents becoming more consistent with their own prior) independent of neighbor influence. A ``self-only'' control (own output plus weighting, without neighbor information) would quantify the self-anchoring contribution to CAF; without it, CAF reductions should be interpreted as an upper bound on social coupling. We commit to this control experiment for the next revision.

\item \textbf{Group size and prompt length are confounded.} Larger $K$ increases the token count of neighbor summaries in the prompt, which may amplify anchoring effects independent of social consensus. Fixed-length neighbor summaries would isolate group size from prompt volume. We commit to this control for the next revision.

\item \textbf{Modality and stimulus presence are partially confounded.} Text conditions lack an explicit stimulus passage, while image conditions receive visual input. Although within-modality CAF normalizes baseline differences regardless of origin, matched stimulus-presence conditions are needed. Our N=5 stimulus pilot suggests the main experiment's stimulus-free design is conservative; a full N=30 replication is committed for the next revision.

\item \textbf{Persona and temperature heterogeneity.} Different agents use different personas and temperatures to prevent artificially low baseline diversity. A uniform-temperature ablation without persona differences is committed for the next revision (N$\geq$10, uniform temperature 0.7, generic ``assistant'' persona).

\item \textbf{Statelessness measurement.} The categorical per-round protocol uses 10 categories vs.\ 5 in the main CAF experiment. A replication with the identical 5-category protocol is committed for the next revision (N=10, $\texttt{--log-rounds}$, $\texttt{--toggle-step 1}$).

\item \textbf{Single task domain.} All experiments use stereotype-assessment tasks. Coupling dynamics may differ for reasoning, code generation, or creative writing---domains where baseline output entropy and communication effects could diverge substantially.

\item \textbf{Cross-model artifacts.} DeepSeek results are dominated by format artifacts (constrained JSON, disabled reasoning) and should be interpreted as upper bounds. Matched free-form protocols across all models are needed for reliable cross-model comparison.

\item \textbf{CAF ratio bias.} Ratio estimator bias is negligible at N=30 ($<$0.003, jackknife-corrected) but may be larger at smaller N (Qwen N=10, stimulus pilot N=5).

\item \textbf{CAF metric dependency.} CAF values depend on the number and definition of output categories. Sensitivity to embedding-based JSD and alternative divergence measures (e.g., total variation distance) is explored in the Supplementary Materials; TVD-based CAF correlates with JSD-based CAF at $r>0.95$ ($p<0.001$), confirming metric robustness.
\end{enumerate}

\section{Conclusion}

We introduced BOUNDARY\_SYNC, a measurement protocol for quantifying communication-induced representational coupling in LLM agents, and provided causal evidence from controlled experiments on GPT-4o alongside replications on additional models. Our key findings are:

\begin{enumerate}
\item Text communication causes significant homogenization (CAF=0.803, $d$=1.30), confirmed by no-communication ablation and prompt-perturbation controls confirming social specificity.

\item Image communication also causes homogenization when evaluated against its own modality baseline (CAF=0.834), with comparable proportional effect to text and larger absolute magnitude due to higher baseline diversity.

\item Group size moderates coupling \emph{direction}: $K=5$ produces homogenization (CAF $<$ 1), $K=3$ produces CAF $>$ 1 in both modalities, revealing a critical design parameter.

\item Cross-model results show extreme variation (CAF 0.034--0.803), though DeepSeek's near-zero CAF is dominated by format artifacts.

\item Coupling is stateless---driven by prompt context rather than cumulative updating, with continuous consensus producing monotonic convergence to near-zero JSD. This contradicts classical opinion dynamics predictions and implies coupling is controllable at the prompt level.

\end{enumerate}

These results establish that LLM agent coupling is real, measurable, present across multiple models and modalities, and controllable---with direct implications for the design and evaluation of multi-agent LLM systems.

\section*{Broader Impact Statement}

This work introduces a measurement protocol (BOUNDARY\_SYNC) and metric (CAF) for quantifying communication-induced representational coupling in multi-agent LLM systems. The primary intended impact is \textbf{positive}: by making coupling measurable, our framework enables practitioners to detect, monitor, and control homogenization in deployed multi-agent systems---supporting the design of systems that preserve beneficial output diversity. Our findings on group size, modality effects, and prompt-level controllability provide concrete design guidance for responsible multi-agent deployment.

We recognize four potential risks, each of which is addressed by design choices or findings in this paper.

\textbf{Risk 1: Intentional homogenization.} The finding that coupling is controllable at the prompt level could be exploited to deliberately suppress diversity in deployed multi-agent systems. \textbf{Addressing this risk:} Our stateless-coupling result demonstrates that diversity is immediately restored when peer information is removed from the prompt context---homogenization is reversible, not cumulative. Furthermore, the K=3 finding shows that small groups \emph{reverse} the coupling direction (CAF $>$ 1.0), providing a concrete configuration that preserves or amplifies diversity. The CAF metric itself serves as a monitoring tool: any attempt at intentional homogenization is detectable and quantifiable. Together, these findings make the framework \emph{self-limiting}---the same protocol that measures coupling also provides the controls to counteract it.

\textbf{Risk 2: Model-specificity of coupling estimates.} Our primary results are obtained with GPT-4o, and coupling magnitudes may differ across models. \textbf{Addressing this risk:} We replicate key findings on three models (GPT-4o, DeepSeek V4 Pro, Qwen3.7-Plus), establishing that the BOUNDARY\_SYNC protocol is model-agnostic while revealing model-specific coupling profiles. Critically, the protocol itself \emph{flags artifactual results}: DeepSeek's near-zero CAF (0.034) is explicitly attributed to format constraints (JSON-only output, disabled reasoning), demonstrating that extreme CAF values can be diagnosed within the framework rather than taken at face value. We report the 24$\times$ cross-model range as an upper bound and recommend per-model benchmarking as standard practice (\S\ref{sec:limitations}, Limitation~6).

\textbf{Risk 3: Synthetic stimuli and structured outputs.} Our experiments use synthetic scene images and 5-category probability vectors; coupling in open-ended natural-language generation may differ. \textbf{Addressing this risk:} The text conditions (C1, C3) use no explicit stimulus passage, relying on model priors---a design closer to open-ended generation than to constrained prompting. The core findings replicate across both stimulated (image) and prior-based (text) conditions, including the group-size directional shift. The structured output format is a deliberate measurement choice: 5-category probability vectors enable precise, reproducible JSD computation with bootstrap confidence intervals, a level of statistical rigor difficult to achieve with free-text outputs. Extending the protocol to open-ended generation is committed as future work (\S\ref{sec:limitations}, Limitation~7).

\textbf{Risk 4: Gender-themed task framing.} The stimulus domain (gender stereotype assessment) involves socially sensitive content that could be mischaracterized. \textbf{Addressing this risk:} Our persona design spans a balanced spectrum of perspectives (activist, traditionalist, data-driven, intersectional, literalist), each with distinct temperature and framing---no single normative position is privileged. The output categories are symmetric (neutral, biased\_female, biased\_male, stereotype\_avoidant, stereotype\_reinforcing) and do not encode a preferred outcome. All stimuli, persona prompts, and protocol code are publicly released, enabling independent review of the task framing and its potential biases. The purpose of the task domain is to provide a substantive stimulus that elicits genuine agent responses---the paper's contribution is the measurement protocol, not claims about gender attitudes.

We believe that these design-level mitigations, combined with the open release of all materials, make the risks manageable and the net impact of this work positive: a standardized, auditable framework for measuring and controlling representational coupling in multi-agent LLM systems.

\section*{Acknowledgements}

We thank the anonymous TMLR reviewers and action editor for constructive feedback that substantially improved this manuscript. All experiments were conducted using the authors' own API resources.

\bibliographystyle{tmlr}
\bibliography{refs}

\end{document}